\documentclass[runningheads]{llncs}
\usepackage[T1]{fontenc}
\usepackage{graphicx}
\usepackage{longtable}
\usepackage{multirow}
\usepackage{booktabs}
\usepackage{amsmath}
\usepackage{indentfirst}
\usepackage{lineno}
\usepackage{float}
\usepackage{cancel}
\usepackage{adjustbox}
\usepackage{tabularx}
\usepackage{lipsum}
\usepackage{amssymb} 
\usepackage{xcolor}  
\usepackage{multirow}
\usepackage{bbding}
\usepackage[normalem]{ulem}
\useunder{\uline}{\ul}{}
\begin{document}
\title{FGENet: Fine-Grained Extraction Network for Congested Crowd Counting}
%
%
\author{Hao-Yuan Ma \and
Li Zhang\inst{(}\Envelope\inst{)} \and
Xiang-Yi Wei}
%
\authorrunning{Ma et al.}
%
\institute{School of Computer Science and Technology, Soochow University, Suzhou, 215006, China \\
\email{zhangliml@suda.edu.cn} 
}
 \maketitle              
\begin{abstract}

Crowd counting has gained significant popularity due to its practical applications. However, mainstream counting methods ignore precise individual localization and suffer from annotation noise because of counting from estimating density maps. Additionally, they also struggle with high-density images.
To address these issues, we propose an end-to-end model called Fine-Grained Extraction Network (FGENet). 
Different from methods estimating density maps, FGENet directly learns the original coordinate points that represent the precise localization of individuals.
This study designs a fusion module, named Fine-Grained Feature Pyramid (FGFP), that is used to fuse feature maps extracted by the backbone of FGENet. 
The fused features are then passed to both regression and classification heads, where the former provides predicted point coordinates for a given image, and the latter determines the confidence level for each predicted point being an individual. 
At the end, FGENet establishes correspondences between prediction points and ground truth points by employing the Hungarian algorithm. 
For training FGENet, we design a robust loss function, named Three-Task Combination (TTC), to mitigate the impact of annotation noise. Extensive experiments are conducted on four widely used crowd counting datasets.
Experimental results demonstrate the effectiveness of FGENet. Notably, our method achieves a remarkable improvement of 3.14 points in Mean Absolute Error (MAE) on the ShanghaiTech Part A dataset, showcasing its superiority over the existing state-of-the-art methods. Even more impressively, FGENet surpasses previous benchmarks on the UCF\_CC\_50 dataset with an astounding enhancement of 30.16 points in MAE. 
\end{abstract}
\section{Introduction}
Crowd counting, a fundamental task in computer vision, aims to accurately estimate the number of individuals in an image. Crowd counting has found practical applications in various fields, such as crowd monitoring and counting customer flow \cite{use1}.
At present, crowd counting methods are all based on deep learning, and most of them count individuals for given images from estimating density maps. This type of methods is under the density-map framework and involves converting an input image into a density map, achieved by smoothing the centroids using multiple Gaussian kernels and subsequently predicting density maps using convolutional neural network (CNN).

\begin{figure}[t]
	\begin{center}
		\includegraphics[width=0.8\textwidth]{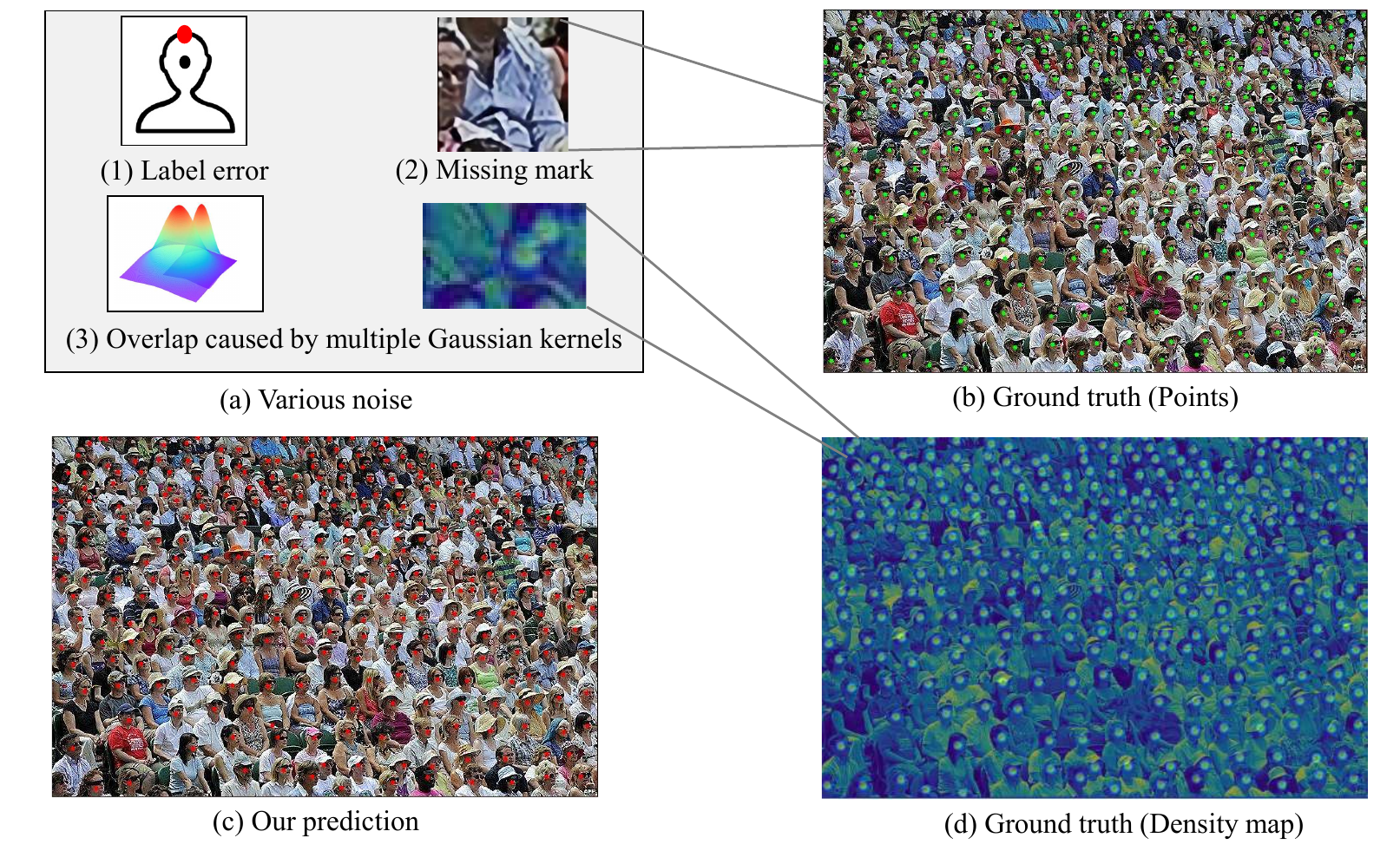}
		\caption{An example from the ShangHaiTech PartA dataset, (a) various noise caused by data annotation ((1) label noise, (2) missing annotation, and (3) overlapping effect caused by Gaussian kernels), (b) the ground truth under the point framework, (c) the prediction map generated by FGENet, and (d) the ground truth under the density-map framework.}\label{fig:01}
	\end{center}
\end{figure}

We can further divide these methods under the density-map framework into two groups: multi-branch \cite{MCNN,mdcnn,multitask} and single-branch \cite{CSRNet,CANNet}.
Although these approaches have made a certain success, they come with a significant drawback of losing the precise location information pertaining to each individual. 
In addition, the superposition effect of Gaussian kernels exacerbates the noise during the annotation process, particularly in high-density scenarios \cite{GauNet}, as vividly illustrated in Figures \ref{fig:01}(a) and \ref{fig:01}(d).   
\par
To remedy above issues, a kind of point framework-based approaches has been proposed \cite{P2PNet,CLTR}. By employing the point framework, models can effectively circumvent the noise generated during the generation of density map and crucially retain the precise location information in the original data. 
Although methods under the point framework can solve the issues caused by the density-map framework, they cannot avoid the noise issue introduced in the annotation process. Both label noise and missing mark (Figure \ref{fig:01}(a)) can ruin the quality of ground truth (Figure \ref{fig:01}(b)) \cite{CHS-Net}. In other words, this issue can introduce inaccuracies and inconsistencies in the ground truth data, impeding the accurate estimation of crowd counting.
Moreover, the preservation of fine-grained information remains a critical concern \cite{CFP}. 
Existing methods often struggle to retain and effectively utilize intricate details in high-density crowd scenarios, limiting their accuracy and precision. 

In a nutshell, crowd counting faces with three challenges at present. (1) Methods based on the density-map framework miss the precise localization of individuals during the counting process. (2) Noise, including label noise, missing mark, and superposition effect of Gaussian kernels, is inevitable when annotating data manually. (3) Existing methods are unsatisfactory in counting high-density crowd images because they may not make full use of fine-grained information. 

To address these challenges, we present an end-to-end crowd counting model, named Fine-Grained Extraction Network (FGENet), that uses the point framework to learn the relationship between original images and coordinate points.
For FGENet, we design a Fine-Grained Feature Pyramid (FGFP) module and an entirely novel Three-Task Combination (TTC) loss
function. FGFP can adeptly capture global dependencies within top-level features and preserve crucial information pertaining to corner regions, while TTC amalgamates the outcomes of classification, regression, and counting tasks. Specially, the TTC loss function could partially mitigate the impact of noise caused by data annotation. 
Figure \ref{fig:01} gives an example from the ShangHaiTech Part\_A dataset. Noise caused by data annotation would degrade the qualification of ground truth under both the point framework and the density-map framework. Figure \ref{fig:01}(d) gives the prediction result of FGENet, which indicates that FGENet is robustness. 
In summary, our contributions are mainly three folds.
\begin{itemize}
    \item  We design a new module, FGFP, that not only enhances the preservation of fine-grained information in feature maps within the same layer but also facilitates effective information fusion of feature maps across different layers.     
    By leveraging the FGFP module, we can overcome the limitations of existing methods that often struggle to capture and integrate intricate details crucial for accurate crowd counting. 
    Through the integration of fine-grained information and the seamless combination of information across feature map layers, our proposed FGFP module serves as a key building block for achieving high counting performance in high-density crowd scenarios.  
    \item We design the TTC loss function that combines classification, regression, and counting tasks, enhancing the generalization ability and robustness of the model. This novel loss function overcomes limitations of traditional approaches and mitigates the impact of training data noise, resulting in a more precise and resilient model that outperforms existing methods. 
    \item On the basis of FGFP and TTC, we propose FGENet, an end-to-end counting model architecture. FGENet can preserve fine-grained information and enhance the accuracy of target counting in high-density scenarios. 
     Extensive experiments unequivocally show that FGENet achieves state-of-the-art (SOTA) performance on diverse high-density datasets, surpassing existing benchmarks. 
\end{itemize}

\section{Related Work}
In this section, we briefly review mainstream crowd counting methods and meticulously examine their respective strengths and weaknesses. 

\subsection{Methods under density-map framework}
In the domain of crowd counting, Lempitsky and Zisserman \cite{2010Learning} introduced the groundbreaking method  that has revolutionized the approach to tackling high-density crowd counting tasks. 
Their approach constructs a linear mapping from local features of an image to a density map, thereby shifting the focus from individual targets to the collective features of target groups. 
As mentioned above, methods based on the density-map framework can be divided into multi-branch and single-branch ones.

\subsubsection{Multi-branch models}
Early counting methods are mostly multi-branch. Some typical multi-branch models are described as follows.  
Multi-column Convolutional Neural Network (MCNN) is a typical multi-branch model that consists of three-column sub-networks \cite{MCNN}. In MCNN, three sub-networks with different convolution kernel sizes are used to extract features of the crowd image separately, and finally the features with three scales are fused by a $1\times 1$ convolution.
Multi-column Mutual Learning (McML) merges a statistical network into a multi-column network to estimate the mutual information between different columns \cite{McML}. 
Dilated-Attention-Deformable ConvNet (DADNet) uses dilated CNNs with different dilation rates to capture more contextual information as the front end and adaptive deformable convolution as the back end to accurately localize the location of objects \cite{DADNet}. 

Above methods are multi-branch but involve in only one task, or density-map estimation. There are multi-task multi-branch models that deal with not only density-map estimation but also other related tasks, for example, crowd count classification \cite{CMTL}, crowd density classification \cite{multitask}, and crowd counting and localization \cite{CL}. 
However, multi-branch models have a significant drawback that they may contain more redundant parameters, leading to reduced efficiency.


\subsubsection{Single-branch models}
Recently, single-branch models become the primary counting methods, and have been proposed for making up the shortcoming of existing models.  
For example, Li et al. \cite{CSRNet} proposed a Congested Scene Recognition Network (CSRNet), Liu et al. \cite{CANNet} proposed Context-Aware Feature Network (CAN), and Ma et al. \cite{FusionCount} proposed the FusionCount. These methods aim at expanding the receptive field of models using different schemes. 

In addition to the issue of receptive field, it is also important to eliminate the effect of backgrounds and solve the problem of extracting characters at different sizes. Thus, Miao et al. \cite{SDANet} proposed a Shallow Feature Based Dense Attention Network (SDANet), which is to weight fusion of multi-level features to obtain the final counting results.
Moreover, researchers have noticed the problem of different distributions in images. Therefore, attention mechanisms have been introduced to solve this problem, such as Attention Scaling NetWork (ASNet) \cite{ASNet} and Multifaceted Attention Network (MAN) \cite{MAN}. 

After that, it was gradually found that the noise of the data has some influence on the final model. 
Thus, researchers have provided different schemes to eliminate noise in data \cite{GLoss,S3,SUA,GauNet,CHS-Net}. For example, Wan et al. \cite{GLoss} presented a Generalized Loss (GLoss) function that can smooth images and serve to eliminate data noise. 
Cheng et al. \cite{GauNet} proposed a Gaussian kernels Network (GauNet), which effectively overcomes the annotated noise in the process of generating density map by replacing the original convolution kernel with a Gaussian kernel. Dai et al. \cite{CHS-Net} proposes a Cross-head Supervision Network (CHS-Net) that solves the noise annotation problem to some extent by applying mutual supervision of convolution head and transformation head.
Semi-balanced Sinkhorn with Scale consistency (S3) changes the prediction objects from density maps to the ground truth values of scatter labeling and the centroids of small Gaussian kernels separately \cite{S3}.

\subsection{Methods under the point framework}
To address the issue of data noise, methods under the point framework have been proposed. These methods directly predict the point locations of individuals instead of estimating density maps. Although S3 operates under the density-map framework, it essentially to estimate point objects generated from density maps.  

Song et al. \cite{P2PNet} proposed a Point to Point Network (P2PNet) that is the first crowd counting method under the point framework. 
P2PNet directly uses the original point coordinates as training data and retains the position information of the individuals during the counting process. After that, a Crowd Localization TRansformer (CLTR) was proposed \cite{CLTR}. 

These methods represent a notable advancement, but they neglect critical factors, such as data noise, intricate feature patterns within feature layers, and corner information—elements that play a pivotal role in accurately addressing the challenges inherent in the high-density counting task \cite{CFP}.

\section{Our Method}
In this section, we present a comprehensive overview of FGENet, outlining the design flow of our overall network, the innovative FGFP module, and the development of the TTC loss function. 

\subsection{Network Design}
\begin{figure}[t]
\begin{center}
\includegraphics[width=0.9\textwidth]{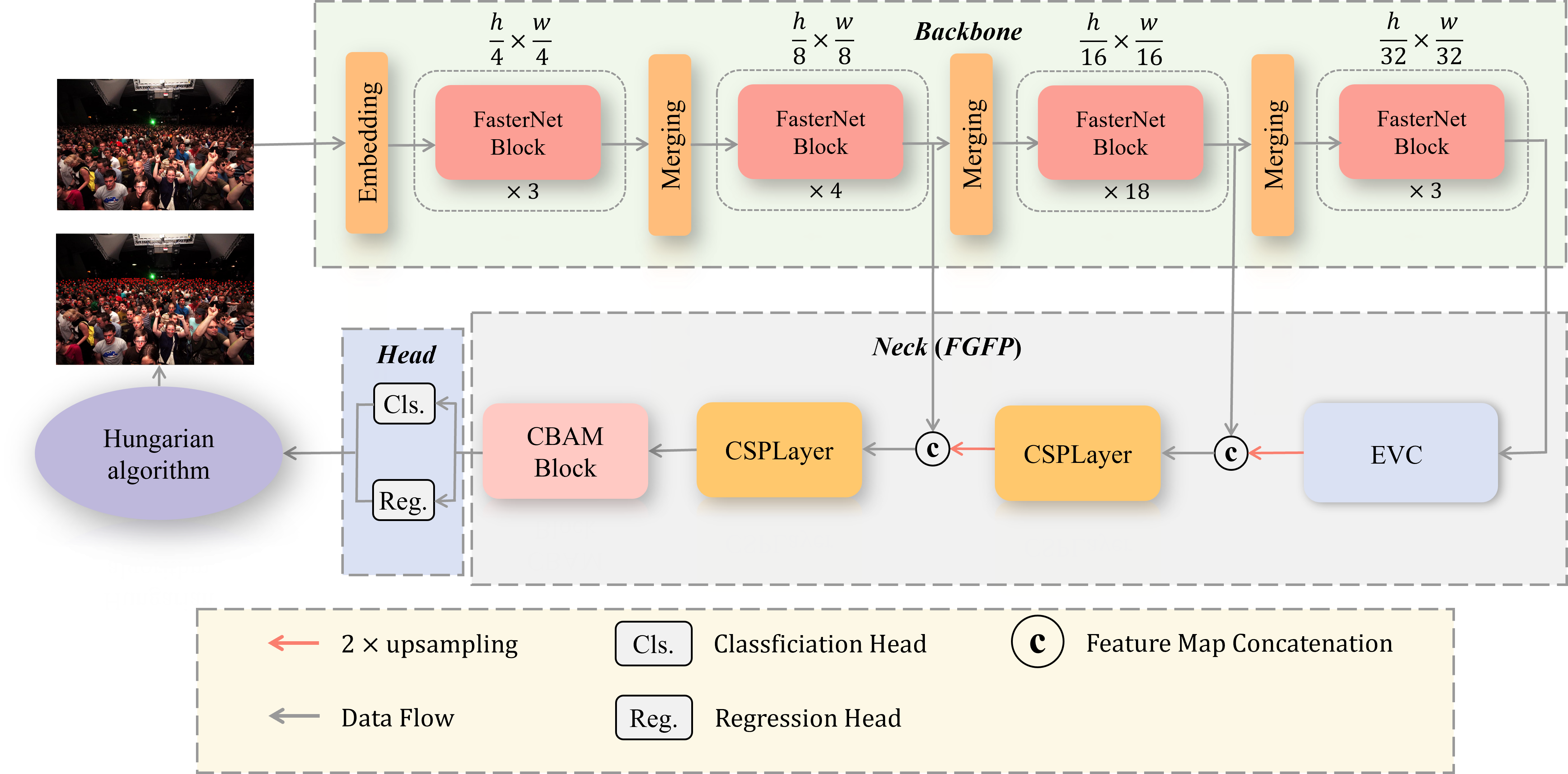}
\caption{Structure of the FGENet.}\label{fig:02}
\end{center}
\end{figure}

The overall architecture of our model, as depicted in Figure \ref{fig:02}, showcases our innovative pipeline design. 
FGENet mainly includes three parts: a backbone, a neck, and a head, where the neck is our designed FGFP module. 
First, we need to select a proper backbone. 
As a general backbone, FasterNet proposed in \cite{FasterNet} has a high accuracy and operation speed because it utilizes the Partial Convolution (PConv). 
Therefore, we leverage the power of FasterNet-L and take it as the backbone of FGENet. 
Subsequently, the extracted features undergo fusion and enhancement through our meticulously designed FGFP module, a critical component that will be extensively discussed in Section 3.2. By leveraging the collective intelligence of these modules, our model accurately predicts the positions of targets and their corresponding confidence levels under the help of dedicated classification and regression tasks. 

Following the aforementioned three parts, FGENet generates a set of precise prediction points, which serve as the foundation for the final result. 
Without loss of generality, let $P$ and $\hat{P}$ be the ordered sets of of coordinate pairs for Ground Truth (GT) and predicted points, respectively, where $P=\{\mathbf{p}_i\}_{i=1}^{N}$,  $\hat{P}=\{\hat{\mathbf{p}}_j\}_{j=1}^{M}$, $N\leq M$, $N$ is the number of GT points, and $M$ is the number of predicted points. 
To ensure accurate correspondence between predicted points and GT points, we employ the Hungarian matching algorithm presented by Kuhn et al. \cite{kuhn1955hungarian}, leveraging a cost matrix as the basis for pairing.
Let $\mathbf{C}\in\mathbb{R}^{N\times M}$ be the cost matrix.
Then, the $i$th row and $j$th column of $C_{ij}$ is calculated as follows:
\begin{align}\label{eq1:cost matrix}
	C_{ij}=\gamma||\mathbf{p}_i-\hat{\mathbf{p}}_j||_2-\hat{t}_j,  i =1,\cdots,N, j=1,\cdots,M,
\end{align}
where $\gamma$ is the equilibrium parameter, $\|\cdot\|_2$ denotes the $L_2$-norm Euclidean distance, and $0\leq \hat{t}_j\leq 1$ is the confidence that the point $\hat{\mathbf{p}}_j$ is predicted as an individual.    
Let $\overline{P}=h(P,\hat{P},\mathbf{C})$, where $h(\cdot)$ is the Hungarian matching algorithm and $\overline{P}=\{\overline{\mathbf{p}}_j\}_{j=1}^M$ is the matching result. In the ordered set $\overline{P}$, the first $N$ points are matched with the corresponding $N$ points in $P$, and the last $(M-N)$ points are lack of matching. 
Through this meticulous pairing process, our method establishes reliable associations between the predicted and GT points, culminating in a robust and accurate prediction outcome.

Finally, the obtained results undergo a critical evaluation through our meticulously designed loss function, which is comprehensively elucidated in Section 3.3. This innovative TTC loss function, tailored to our specific problem domain, serves as a fundamental component in the optimization of our model parameters. Leveraging back-propagation, we iteratively refine and update the model parameters to maximize the model performance and enhance the overall accuracy and robustness of our approach.

\subsection{FGFP Module}
FGFP is a fine-grained extraction module, this module is capable of preserving the fine-grained information in the same layer of the feature map and fusing the valid information contained in different layers of the feature map. 

\par
The $1/32$ feature map is initially processed by the Explicit Visual Center (EVC) block proposed in \cite{CFP}. Subsequently, the information is up-sampled from the input module and fused with the $1/16$ feature map. This process is repeated for the $1/8$ feature map, followed by the application of the Convolutional Block Attention Module (CBAM) to enhance overall performance.

\subsection{TTC Loss}
Once the predicted points have been matched by using the Hungarian matching algorithm, we proceed to employ a meticulously designed TTC loss to guide the training process. TTC encompasses three distinct components: regression loss, classification loss, and count loss, each addressing specific aspects of our crowd counting task. 

To mitigate the effect of noise in data, we design a novel regression loss $L_{reg}$, called highly smoothed $L_1$ (HSL$_1$) loss. HSL$_1$ is a variant of the smooth $L_1$ loss and can effectively minimize the influence of data noise on the final model, which has the form
\begin{align}\label{eq3:loc loss}
L_{reg}=\frac{1}{N}\sum\limits_{i=1}^{N}log(\textit{s}(\mathbf{p}_i,\overline{\mathbf{p}}_i)+1),
\end{align}
where the $L_1$ smoothing loss $s(\cdot,\cdot)$ can be expressed as
\begin{align}\label{eq2:smooth l1}
s(\mathbf{p}_i,\overline{\mathbf{p}}_i),
=\begin{cases}
0.5\|\mathbf{p}_i-\overline{\mathbf{p}}_i\|^2_2,\text{ if } \|\mathbf{p}_i-\overline{\mathbf{p}}_i\|_1<1 
\\
\|\mathbf{p}_i-\overline{\mathbf{p}}_i\|_1-0.5,\text{ otherwise}
\end{cases}
\end{align}
and $\|\cdot\|_1$ denotes the $L_1$-norm. 

Regarding the classification loss $L_{cls}$, we adapt the Cross Entropy (CE) loss proposed in \cite{csloss} by introducing a carefully calibrated weight $\alpha$ and subsequently propose a Weighted CE (WCE) loss. Here, WCE for classification is defined as
\begin{align}\label{eq4:cls loss}
L_{cls}=-\frac{1}{N} \left\{\alpha\sum\limits_{i=1}^{N}\log\overline{t}_i+(1-\alpha)\sum\limits_{i=N+1}^{M}\left(1-\log\overline{t}_{i}\right)\right\},
\end{align}
where $\overline{t}_i$ is the confidence that $\overline{\mathbf{p}}_i$ is predicted as an individual, and the weight $\alpha\in [0,1]$. 

For the counting task, we design a Highly Robust Count (HRC) loss, that is 
\begin{align}\label{eq5:num loss}
L_{cou}=|M-N|\log\left(\frac{|M-N|}{N+\epsilon}+1\right),
\end{align}
where $\epsilon$ is a very small positive constant. The HRC loss effectively emphasizes errors in high-density counts and outperforms the traditional Mean Absolute Error (MAE), as demonstrated in our experiments. 

Finally, the overall loss function $L$ is defined as a weighted combination of the three aforementioned components, contributing to the fine-tuning of the model performance. The comprehensive formulation of the loss function is expressed as
\begin{align}\label{eq6:loss sum}
L=\lambda_1 L_{cls}+\lambda_2 L_{reg}+\lambda_3 L_{cou},
\end{align}
where $\lambda_1$, $\lambda_2$, and $\lambda_3$ are the weights for combining three loss components.

\section{Experiments}
In this section, we embark on a comprehensive exploration of our proposed approach through a series of rigorous experiments. Our algorithm is coded in Python, using the Pytorch deep learning framework. Experiments are performed on an Ubuntu 18.04 system with a GeForce RTX 3090 graphics card.
\subsection{Datasets}
In our experiments, there are four public datasets, including ShangHaiTech Part\_A and Part\_B \cite{MCNN}, UCF\_CC\_50 \cite{UCFCC50}, and UCF-QNRF \cite{QNRF}, as shown in Table \ref{tab:1}, where SHT\_A is short for ShangHaiTech Part\_A, and SHT\_B is for ShangHaiTech Part\_B. The description about these datasets are as follows
\begin{itemize}
	\item \textbf{SHT\_A:} This dataset is from the ShangHaiTech dataset that holds immense significance in crowd counting. SHT\_A is regarded as a cornerstone dataset and has earned its reputation as one of the largest and most widely used datasets presently.  
	This dataset encompasses a comprehensive collection of 482 web images, with a division of 300 images for training and 182 images for rigorous test. In addition, SHT\_A contains an extensive collection of marker points, totaling 241,677 individuals. 
	On average, there are approximately 501.4 marker points per image, highlighting the rich and detailed annotations present in the dataset. 
	
	\item \textbf{SHT\_B:} This dataset is also from the ShangHaiTech dataset. 
 SHT\_B offers a unique perspective with busy street images captured across Shanghai. With a total of 716 images, there are 400 images dedicated to training and 316 images for test. 
  This dataset presents diverse scene types, encompassing varied perspective angles and crowd density, thus providing a rich and representative collection for comprehensive evaluation.

	\item \textbf{UCF\_CC\_50:} This dataset presents a unique challenge in crowd counting research. With a total of 50 images, this dataset offers a diverse range of scenes, including concerts, protests, stadiums, and marathons.
     Each image encapsulates diverse crowd densities and perspectives. Crowd density in the UCF\_CC\_50 dataset range from 94 to 4543 individuals.
   This dataset serves as a valuable resource for evaluating the performance and accuracy of crowd counting methods across challenging and diverse real-world scenarios.
	
	\item \textbf{UCF-QNRF:} This dataset is a comprehensive and extensive collection that significantly contributes to the field of crowd counting research. With a staggering 1535 images, this dataset provides a rich and diverse set of data for analysis. The dataset is derived from web images and encompasses a massive 1,251,642 tags, serving as ground truth annotations for individual targets.
\end{itemize}

\begin{table}[]
\centering
\scriptsize  
\caption{Four crowd counting datasets used in experiments.}\label{tab:1}
\begin{tabular}{@{}lccccccc@{}}
\toprule
\multicolumn{1}{l}{\multirow{2}{*}{Dataset}} & \multirow{2}{*}{\#Images~~} & \multirow{2}{*}{\#Training/Test~~} & \multicolumn{5}{c}{Count statistics}                                        \\ \cmidrule(l){4-8} 
\multicolumn{1}{c}{}                         &                                   &                              & \multicolumn{1}{l}{Average resolution} & Total     & Min~ & Average~ & Max    \\ \midrule
SHT\_A                                       & 482                               & 300/182                      & 589$\times$868                              & 241,677   & 33  & 501.4   & 3139  \\ 
SHT\_B                                       & 716                               & 400/316                      & 768$\times$1024                             & 88,488    & 9   & 123.6   & 578    \\ 
UCF\_CC\_50                                  & 50                                & -                            & 2101$\times$2888                            & 63,974    & 94  & 1280   & 4543  \\ 
UCF-QNRF                                     & 1535                             & 1201/334                     & 2013$\times$2902                            & 1,251,642 & 49  & 815     & 12,865 \\ \bottomrule
\end{tabular}
\end{table}



\subsection{Model Evaluation}
To benchmark our approach, we compare it against 16 methods, including MCNN \cite{MCNN}, CSRNet \cite{CSRNet}, CAN \cite{CANNet}, SDANet \cite{SDANet}, ASNet \cite{ASNet}, GLoss \cite{GLoss}, S3 \cite{S3}, Spatial Uncertainty-Aware (SUA) \cite{SUA}, P2PNet \cite{P2PNet}, FusionCount \cite{FusionCount}, MAN \cite{MAN}, GauNet \cite{GauNet}, Characteristic Function Loss (ChfL) \cite{ChfL}, CLTR \cite{CLTR}, OrdinalEntropy \cite{OrdinalEntropy}, and CHS-Net \cite{CHS-Net}.
The evaluation metrics employed for assessing the model performance are MAE and Mean Squared Error (MSE), which are widely adopted in crowd counting.

\begin{table}[t]
\setlength{\abovecaptionskip}{0cm}  
	\setlength{\belowcaptionskip}{10pt} 
\caption{Comparison of counting performance obtained by 17 methods.}	\label{tab:3}
	\scriptsize 
\begin{tabular}{@{}llcccccccc@{}}
\toprule
\multicolumn{1}{l}{\multirow{2}{*}{Methods}} & \multicolumn{1}{l}{\multirow{2}{*}{Venue}} & \multicolumn{2}{c}{SHT\_A}                  & \multicolumn{2}{c}{SHT\_B}                 & \multicolumn{2}{c}{UCF\_CC\_50}                   & \multicolumn{2}{c}{UCF-QNRF}                      \\ \cmidrule(l){3-10} 
\multicolumn{1}{c}{}                         & \multicolumn{1}{c}{}                       & \multicolumn{1}{l}{MAE}  & \multicolumn{1}{l}{MSE} & \multicolumn{1}{l}{MAE} & \multicolumn{1}{l}{MSE} & \multicolumn{1}{l}{MAE} & \multicolumn{1}{l}{MSE} & \multicolumn{1}{l}{MAE} & \multicolumn{1}{l}{MSE} \\ \midrule
MCNN \cite{MCNN}                                         & CVPR'16                                    & 110.2                    & -                       & 26.4                    & -                       & 377.6                   & -                       & -                       & -                       \\ 
CSRNet \cite{CSRNet}                                        & CVPR'18                                    & 68.2                     & 115                     & 10.6                    & 16                      & 266.1                   & -                       & 120.3                   & 208.5                   \\ 
CAN \cite{CANNet}                                          & CVPR'19                                    & 62.8                     & 101.8                   & 7.7                     & 12.7                    & 212.2                   & 243.7                   & 107                     & 183                     \\ SDANet \cite{SDANet}                                       & AAAI'20                                    & 63.6                     & 101.8                   & 7.8                     & {\ul 10.2}                    & 227.6                   & 316.4                   & -                       & -                       \\ ASNet \cite{ASNet}                                        & CVPR'20                                    & 57.78                    & 90.13                   & -                       & -                       & 174.84                  & {\ul 251.63}                  & 91.59                   & 159.71                  \\ 
GLoss \cite{GLoss}                                        & IJCAI'21                                   & 57.3                     & 90.7                    & 7.3                     & 11.4                    & -                       & -                       & 81.2                    & 138.6                   \\ 
S3 \cite{S3}                                           & IJCAI'21                                   & 57                       & 96                      & {\ul 6.3}                     & 10.6                    & -                       & -                       & 80.6                    & 139.8                   \\ 
SUA \cite{SUA}                                          & ICCV'21                                    & 68.5                     & 121.9                   & 14.1                    & 20.6                    & -                       & -                       & 130.3                   & 226.3                   \\ 
P2PNet \cite{P2PNet} & ICCV'21 & 55.73 & 89.20 & 6.97 & 11.34 & {\ul 172.72} & 256.18 & 85.32 & 154.5 \\ 

FusionCount \cite{FusionCount}                                  & ICIP'22                                    & 62.2                     & 101.2                   & 6.9                     & 11.8                    & -                       & -                       & -                       & -                       \\ 
MAN \cite{MAN}                                          & CVPR'22                                    & 56.8                     & 90.3                    & -                       & -                       & -                       & -                       & \textbf{77.3}              & \textbf{131.5}             \\ 
GauNet \cite{GauNet}                                       & CVPR'22                                    & {\ul 54.8}                     & {\ul 89.1}                    & \textbf{6.2}               & \textbf{9.9}               & 186.3                   & 256.5                   & 81.6                    & 153.7                   \\ 
ChfL \cite{ChfL}                                         & CVPR'22 & 57.5 & 94.3                    & 6.9                     & 11               & -                       & -                       & {\ul 80.3}                    &{\ul 137.6}                   \\ 
CLTR \cite{CLTR}                                         & ECCV'23                                    & 56.9                     & 95.2                    & 6.5                     & 10.6                    & -                       & -                       & 85.8                    & 141.3                   \\ 
OrdinalEntropy \cite{OrdinalEntropy}                               & ICLR'23                                    & 65.6                     & 105                     & 9.1                     & 14.5                    & -                       & -                       & -                       & -                       \\ 
CHS-Net \cite{CHS-Net}                                      & ICASSP'23                                  & 59.2                     & 97.8                    & 7.1                     & 12.1                    & -                       & -                       & 83.4                    & 144.9                   \\ \midrule
FGENet (Ours)                                         & \multicolumn{1}{c}{-}                      & \textbf{51.66}           & \textbf{85}             & 6.34                    & {10.53}                   & \textbf{142.56}         & \textbf{215.87}         & 85.2                    & 158.76                  \\ \bottomrule
\end{tabular}
\end{table}

\begin{figure}[t]
	\begin{center}
		\vspace{0pt}
 
		\setlength{\belowdisplayskip}{0pt} 
		\includegraphics[width=0.95\textwidth]{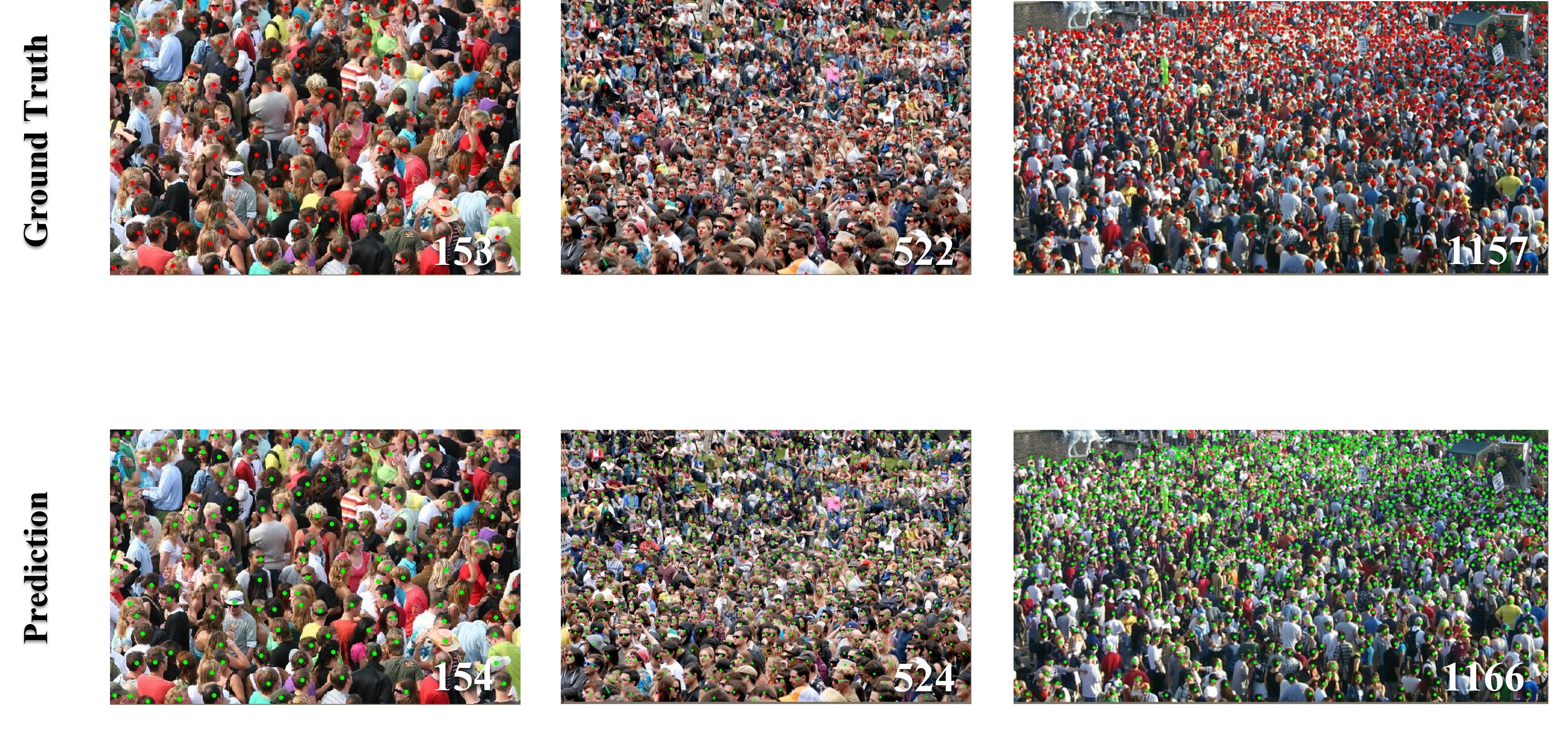}
		\caption{Three crowd images of SHT\_A and their predictions obtained by FGENet.}\label{fig:03}
	\end{center}
\end{figure}

Table \ref{tab:3} lists the comparison of FGENet with other 16 methods for crowd counting, where the best results are in bold and the second best ones are underlined. Observation on Table \ref{tab:3} indicates that FGENet achieves the SOTA results on both SHT\_A and UCF\_CC\_50.
Specifically, FGENet significantly improves 5.72\% MAE on SHT\_A, and 17.46 \% MAE on the UCF\_CC\_50 dataset compared to the second best method. 
In Figure \ref{fig:03}, three images from SHT\_A are given, and their predictions obtained by FGENet are also provided. We can see the number of predicted points is very close to the number of GT points. 

On SHT\_B, FGENet ranks the third but is only 0.14 MAE less than the best model. As indicated in Table \ref{tab:1}, 
there are a large amount of missing annotations in the SHT\_B dataset. Although our method can capture high occlusion small density targets, it is not able to cope with this situation well. 

Due to images with large size in UCF-QNRF, the methods related to Transformer work better, such as MAN. In addition, our method is limited by the efficiency of the matching algorithm, so there are some issues in the process of parameter adjustment.

In summary, we find that the existing models can fit the dataset with large-scale and high-density or small-scale and low-density, but perform generally bad on small-scale and high-density datasets. Fortunately, our method can remedy this defect because experimental results demonstrate that FGENet is particularly effective on small-scale and high-density datasets.

\subsection{Ablation Study}
The goal of ablation study is to validate the efficacy and reliability of new designed FGFP module and TTC loss function. Here, ablation experiments are conducted on the SHT\_A dataset. 
\subsubsection{Experiments on FGFP}
To demonstrate the validity of our FGFP, we list the possible combinations of our model.
The ablation settings and experimental results are shown in Table \ref{tab:4}. 
Among these combinations, our model with the complete FGFP module (ID8 in Table \ref{tab:4}) has the best counting performance, and our model without FGFP (ID1 in Table \ref{tab:4}) is the worst.
Observation on ID2--ID4 in Table \ref{tab:4} indicates that CSPLayer is the most important block among three blocks, and observation on ID5--ID7 shows that the combination of CSPLayer and CBAM is compared to that of CSPLayer and EVC. 

In summary, findings demonstrate the scientific nature of our proposed FGFP module. In FGFP, the goal of CSPLayer is to fuse the information of feature maps between layers. EVC and CBAM are used to extract the fine-grained information within feature maps and enhance attention to different channels and spaces, respectively.  
The three blocks complement each other, and neither is dispensable. 
\begin{table}[h]
	\small
\centering
\setlength{\abovecaptionskip}{0cm}  
\setlength{\belowcaptionskip}{10pt} 
\caption{Ablation experiments on FGFP.}	\label{tab:4}
\begin{tabular}{lccccc}
\hline
 ID~&EVC~~&CSPLayer~~&CBAM~~& MAE~~ & MSE ~~\\
 \hline
1 & × & × & × & 72.19 & 112.18 \\ 
2 & \checkmark & × & × & 71.84 & 112.09 \\ 
3 & × & \checkmark & × & 57.6 & 92.27 \\ 
4 & × & × & \checkmark & 69.51 & 116.6 \\ 
5 & \checkmark & \checkmark & × & 54.67 & 96.99 \\ 
6 & \checkmark & × & \checkmark & 62.79 & 109.84 \\ 
7 & × & \checkmark & \checkmark & 54.1 & 92.44 \\ 
8 & \checkmark & \checkmark & \checkmark & 51.66 & 85 \\ \hline
\end{tabular}
\end{table}

\subsubsection{Experiments on loss function}
The TTC loss function (\ref{eq6:loss sum}) consists of three terms: HSL$_1$ $L_{reg}$ for regression, WCE $L_{cls}$ for classification, and HRC $L_{cou}$ for counting. To validate the performance of TTC loss, we set up 10 possible variants and show their experimental results in Table \ref{tab:5}. Note that ID1 is common used loss, and ID2 is the baseline of our loss. 

Among variants ID2--ID5, we can see that ID4 has the best result, which changes only CE to WCE for classification. The main reason is that the weight $\lambda_1$ of the classification task accounts for a higher weight in the whole loss function. 
In the case of changing two terms (variants ID6--ID8), ID8 using WCE and HRC works better. We believe that the smoothness and robustness of HRC solves the problem caused by the missed mark to a certain extent.
Among variants ID8--ID10, the variant ID10 performs the best. Because HSL$_1$ is less penalized for smaller position errors and overly very smooth, it can reduce the influence of labeling errors.

\begin{table}[h]
\centering
\small
\setlength{\abovecaptionskip}{0cm}  
\setlength{\belowcaptionskip}{10pt} 
\caption{Ablation experiments on TTC loss function.}	\label{tab:5}
\begin{tabular}{lccccc}
\hline
ID~&Regression loss~&Classification loss~ &Counting loss~& MAE~&MSE~\\ 
\hline
1 & MSE & CE & \textbf{-} & 59.62 & 104.12 \\ 
2 & MSE & CE & MAE & 57.18 & 98.21 \\ 
3 & HSL$_1$ & CE & MAE & 57.115 & 97.62 \\ 
4 & MSE & WCE & MAE & 53.37 & 92.72 \\
5 & MSE & CE & HRC & 56.36 & 100.80 \\ 
6 & HSL$_1$ & WCE & MAE & 53.35 & 90.74 \\ 
7 & HSL$_1$ & CE & HRC & 56.59 & 95.02 \\
8 & MSE & WCE & HRC & 52.14 & 87.37 \\ 
9 & Smooth L1 & WCE & HRC & 54.64 & 95.54 \\ 
10 & HSL$_1$ & WCE & HRC & \textbf{51.66} & \textbf{85.00} \\ \hline
\end{tabular}
\end{table}


\section{Conclusion}
In our study, we have shed light on two prevalent challenges in the field of crowd counting. First, we have identified three distinct types of noise that manifest in the training data, which inevitably impact the performance of the final model. Second, given the pixel-level nature of crowd counting, the preservation and utilization of fine-grained information play a crucial role in achieving accurate results. To effectively tackle these challenges, we have introduced a novel loss function to mitigate the influence of noise, and a dedicated FGFP module to address the fine-grained information problem.
Through extensive experimentation, we have demonstrated the scientific validity and efficacy of our proposed approach. 

Our method has achieved state-of-the-art (SOTA) results on benchmark datasets, such as ShangHaiTech PartA and UCF\_CC\_50, showcasing its superior performance compared to existing methods. However, it is important to acknowledge that our method does have certain limitations. For instance, it exhibits relatively lower performance in detecting large targets, and the matching process for high-resolution images incurs high time complexity.
Moving forward, we are committed to further optimizing our detection results and refining our matching algorithm to overcome these limitations. 
\clearpage

%
%
%
%




\bibliographystyle{splncs04}
\bibliography{reference}

\begin{thebibliography}{10}
\providecommand{\url}[1]{\texttt{#1}}
\providecommand{\urlprefix}{URL }
\providecommand{\doi}[1]{https://doi.org/#1}

\bibitem{FasterNet}
Chen, J., Kao, S.h., He, H., Zhuo, W., Wen, S., Lee, C.H., Chan, S.H.G.: {Run, Don't Walk: Chasing Higher FLOPS for Faster Neural Networks}. In: Proceedings of the IEEE/CVF Conference on Computer Vision and Pattern Recognition (CVPR). pp. 12021--12031 (2023)

\bibitem{GauNet}
Cheng, Z.Q., Dai, Q., Li, H., Song, J., Wu, X., Hauptmann, A.G.: {Rethinking Spatial Invariance of Convolutional Networks for Object Counting}. In: Proceedings of the IEEE/CVF Conference on Computer Vision and Pattern Recognition (CVPR). pp. 19638--19648 (2022)

\bibitem{McML}
Cheng, Z.Q., Li, J.X., Dai, Q., Wu, X., He, J.Y., Hauptmann, A.: {Improving the Learning of Multi-column Convolutional Neural Network for Crowd Counting}  (2019)

\bibitem{CHS-Net}
Dai, M., Huang, Z., Gao, J., Shan, H., Zhang, J.: {Cross-Head Supervision for Crowd Counting with Noisy Annotations}. In: Proceedings of the IEEE International Conference on Acoustics, Speech and Signal Processing (ICASSP). pp.~1--5. IEEE (2023)

\bibitem{DADNet}
Guo, D., Li, K., Zha, Z., Wang, M.: {DADNet: Dilated-Attention-Deformable ConvNet for Crowd Counting}. In: Proceedings of the the 27th ACM International Conference (ACM MM) (2019)

\bibitem{UCFCC50}
Idrees, H., Saleemi, I., Seibert, C., Shah, M.: {Multi-source Multi-scale Counting in Extremely Dense Crowd Images}. In: Proceedings of the IEEE Conference on Computer Vision and Pattern Recognition (CVPR). pp. 2547--2554 (2013)

\bibitem{CL}
Idrees, H., Tayyab, M., Athrey, K., Zhang, D., Al-Maadeed, S., Rajpoot, N., Shah, M.: {Composition Loss for Counting, Density Map Estimation and Localization in Dense Crowds}. In: Proceedings of the European Conference on Computer Vision (ECCV) (2018)

\bibitem{QNRF}
Idrees, H., Tayyab, M., Athrey, K., Zhang, D., Al-Maadeed, S., Rajpoot, N., Shah, M.: {Composition Loss for Counting, Density Map Estimation and Localization in Dense Crowds}. In: Proceedings of the European conference on computer vision (ECCV). pp. 532--546 (2018)

\bibitem{ASNet}
Jiang, X., Zhang, L., Xu, M., Zhang, T., Lv, P., Zhou, B., Yang, X., Pang, Y.: {Attention Scaling for Crowd Counting}. In: Proceedings of the IEEE/CVF Conference on Computer Vision and Pattern Recognition (CVPR). pp. 4706--4715 (2020)

\bibitem{kuhn1955hungarian}
Kuhn, H.W.: {The Hungarian Method for the Assignment Problem}. Naval research logistics quarterly  \textbf{2}(1-2),  83--97 (1955)

\bibitem{2010Learning}
Lempitsky, V.S., Zisserman, A.: {Learning To Count Objects in Images}. In: Proceedings of the 24th Annual Conference on Neural Information Processing Systems (NeurIPS) (2010)

\bibitem{CSRNet}
Li, Y., Zhang, X., Chen, D.: {CSRNet: Dilated Convolutional Neural Networks for Understanding the Highly Congested Scenes}. In: Proceedings of the IEEE Conference on Computer Vision and Pattern Recognition (CVPR). pp. 1091--1100 (2018)

\bibitem{CLTR}
Liang, D., Xu, W., Bai, X.: {An End-to-End Transformer Model for Crowd Localization}. In: Proceedings of the European Conference on Computer Vision (ECCV). pp. 38--54 (2022)

\bibitem{S3}
Lin, H., Hong, X., Ma, Z., Wei, X., Qiu, Y., Wang, Y., Gong, Y.: {Direct Measure Matching for Crowd Counting}. In: Proceedings of the Thirtieth International Joint Conference on Artificial Intelligence (IJCAI). pp. 837--844 (2021)

\bibitem{MAN}
Lin, H., Ma, Z., Ji, R., Wang, Y., Hong, X.: {Boosting Crowd Counting via Multifaceted Attention}. In: Proceedings of the IEEE/CVF Conference on Computer Vision and Pattern Recognition (CVPR). pp. 19628--19637 (2022)

\bibitem{CANNet}
Liu, W., Salzmann, M., Fua, P.: {Context-Aware Crowd Counting}. In: Proceedings of the IEEE/CVF Conference on Computer Vision and Pattern Recognition (CVPR). pp. 5099--5108 (2019)

\bibitem{FusionCount}
Ma, Y., Sanchez, V., Guha, T.: {FusionCount: Efficient Crowd Counting via Multiscale Feature Fusion}. In: Proceedings of the IEEE International Conference on Image Processing (ICIP). pp. 3256--3260 (2022)

\bibitem{SUA}
Meng, Y., Zhang, H., Zhao, Y., Yang, X., Qian, X., Huang, X., Zheng, Y.: {Spatial Uncertainty-Aware Semi-Supervised Crowd Counting}. In: Proceedings of the IEEE/CVF International Conference on Computer Vision (ICCV). pp. 15549--15559 (2021)

\bibitem{SDANet}
Miao, Y., Lin, Z., Ding, G., Han, J.: {Shallow Feature Based Dense Attention Network for Crowd Counting}. In: Proceedings of the AAAI Conference on Artificial Intelligence (AAAI). pp. 11765--11772 (2020)

\bibitem{CFP}
Quan, Y., Zhang, D., Zhang, L., Tang, J.: {Centralized Feature Pyramid for Object Detection}. CoRR  \textbf{abs/2210.02093} (2022)

\bibitem{ChfL}
Shu, W., Wan, J., Tan, K.C., Kwong, S., Chan, A.B.: {Crowd Counting in the Frequency Domain}. In: Proceedings of the IEEE/CVF Conference on Computer Vision and Pattern Recognition (CVPR). pp. 19618--19627 (2022)

\bibitem{CMTL}
Sindagi, V.A., Patel, V.M.: {CNN-based Cascaded Multi-task Learning of High-level Prior and Density Estimation for Crowd Counting}. In: Proceedings of the 14th IEEE international conference on advanced video and signal based surveillance (AVSS). pp.~1--6. IEEE (2017)

\bibitem{P2PNet}
Song, Q., Wang, C., Jiang, Z., Wang, Y., Tai, Y., Wang, C., Li, J., Huang, F., Wu, Y.: {Rethinking Counting and Localization in Crowds:A Purely Point-Based Framework}. In: Proceedings of the IEEE/CVF International Conference on Computer Vision (ICCV). pp. 3365--3374 (2021)

\bibitem{GLoss}
Wan, J., Liu, Z., Chan, A.B.: {A Generalized Loss Function for Crowd Counting and Localization}. In: Proceedings of the IEEE/CVF Conference on Computer Vision and Pattern Recognition (CVPR). pp. 1974--1983 (2021)

\bibitem{csloss}
Wang, Y., Ma, X., Chen, Z., Luo, Y., Yi, J., Bailey, J.: {Symmetric Cross Entropy for Robust Learning With Noisy Labels}. In: Proceedings of the IEEE/CVF International Conference on Computer Vision (CVPR). pp. 322--330 (2019)

\bibitem{use1}
Wen, L., Du, D., Zhu, P., Hu, Q., Wang, Q., Bo, L., Lyu, S.: {Detection, Tracking, and Counting Meets Drones in Crowds: A Benchmark}. In: Proceedings of the IEEE/CVF Conference on Computer Vision and Pattern Recognition (CVPR). pp. 7812--7821 (2021)

\bibitem{mdcnn}
Yan, L., Zhang, L., Zheng, X., Li, F.: {Deeper multi-column dilated convolutional network for congested crowd understanding}. Neural Computing and Applications  \textbf{34}(2),  1407--1422 (2022)

\bibitem{multitask}
Zhang, L., Yan, L., Zhang, M., Lu, J.: {T\({}^{\mbox{2}}\){CNN}: a novel method for crowd counting via two-task convolutional neural network}. Visual Comput  \textbf{39}(1),  73--85 (2023)

\bibitem{OrdinalEntropy}
Zhang, S., Yang, L., Mi, M.B., Zheng, X., Yao, A.: {Improving Deep Regression with Ordinal Entropy} (2023), \url{https://doi.org/10.48550/arXiv.2301.08915}

\bibitem{MCNN}
Zhang, Y., Zhou, D., Chen, S., Gao, S., Ma, Y.: {Single-Image Crowd Counting via Multi-Column Convolutional Neural Network}. In: Proceedings of the IEEE Conference on Computer Vision and Pattern Recognition (CVPR). pp. 589--597 (2016)

\end{thebibliography}
\end{document}